\documentclass[10pt,twocolumn,letterpaper]{article}

\usepackage{wacv}
\usepackage{times}
\usepackage{epsfig}
\usepackage{graphicx}
\usepackage{amsmath}
\usepackage{amssymb}
\usepackage{booktabs}
\usepackage{subfig}
\usepackage[inline]{enumitem}
\usepackage{bm}

\newcommand{\myfirstpara}[1]{\par \noindent \textbf{#1:}}
\newcommand{\mypara}[1]{\vspace{0.5em} \myfirstpara{#1}}

\newcommand\SOTA{\texttt{SOTA}\xspace}

\def\wacvPaperID{1730} 

\wacvalgorithmstrack   
\wacvfinalcopy 

\usepackage[pagebackref=true,breaklinks=true,colorlinks,bookmarks=false]{hyperref}

\usepackage[capitalise]{cleveref}

\begin{document}

\title{Attention Attention Everywhere: \\ Monocular Depth Prediction with Skip Attention}
\author{Ashutosh Agarwal \qquad
Chetan Arora 
\\
IIT Delhi\\
{\tt\small \{ashutosh.agarwal,chetan\}@cse.iitd.ac.in}
}

\maketitle
\thispagestyle{empty}

\begin{abstract}

Monocular Depth Estimation (MDE) aims to predict pixel-wise depth given a single RGB image. For both, the convolutional as well as the recent attention-based models, encoder-decoder-based architectures have been found to be useful due to the simultaneous requirement of global context and pixel-level resolution. Typically, a skip connection module is used to fuse the encoder and decoder features, which comprises of feature map concatenation followed by a convolution operation. Inspired by the demonstrated benefits of attention in a multitude of computer vision problems, we propose an attention-based fusion of encoder and decoder features. We pose MDE as a pixel query refinement problem, where coarsest-level encoder features are used to initialize pixel-level queries, which are then refined to higher resolutions by the proposed Skip Attention Module (SAM). We formulate the prediction problem as ordinal regression over the bin centers that discretize the continuous depth range and introduce a Bin Center Predictor (BCP) module that predicts bins at the coarsest level using pixel queries. Apart from the benefit of image adaptive depth binning, the proposed design helps learn improved depth embedding in initial pixel queries via direct supervision from the ground truth. Extensive experiments on the two canonical datasets, NYUV2 and KITTI, show that our architecture outperforms the state-of-the-art by 5.3\% and 3.9\%, respectively, along with an improved generalization performance by 9.4\% on the SUNRGBD dataset.
Code is available at \url{https://github.com/ashutosh1807/PixelFormer.git}.
\end{abstract}

\section{Introduction}
%Recent surge in applications such as autonomous driving, robotics, and 3D reconstruction has sparked an interest among researchers in Monocular Depth Estimation (MDE), which aims to predict pixel-wise depth using a single RGB image as an input. Using a single image to estimate depth is an ill-posed problem due to scale ambiguity because multiple 3D scenes can be back-projected to the same 2D scene. This makes MDE more complex than other scene understanding problems, such as semantic segmentation etc. 

% \begin{figure}[t]
% \includegraphics[width=1.0\linewidth, height=0.4\linewidth]{figures/intro.png}
% \caption{\textbf{Left}: Input RGB image. \textbf{Center} : Depth map predicted by NeWCRF \cite{newcrf}. Right: Depth map predicted by Pixelformer (Ours). We see certain objects missing in NewCRF which gets resolved by Pixelformer via effective encoder-decoder fusion.}
% \label{fig:intro}
% \end{figure}

\begin{figure}
\begin{center}
\setlength\tabcolsep{2pt}
\renewcommand{\arraystretch}{1.1}
\begin{tabular}{ccccc}
\includegraphics[width=0.23\linewidth]{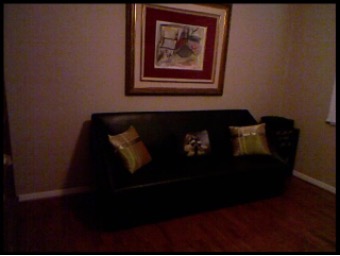} &
\includegraphics[width=0.23\linewidth]{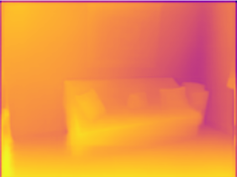} &
\includegraphics[width=0.23\linewidth]{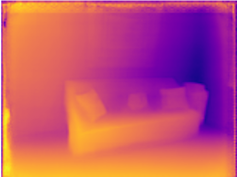} &
\includegraphics[width=0.23\linewidth]{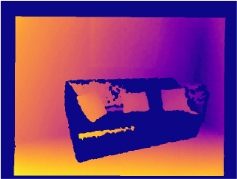} 
\\
\includegraphics[width=0.23\linewidth]{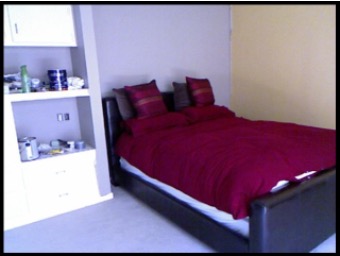} &
\includegraphics[width=0.23\linewidth]{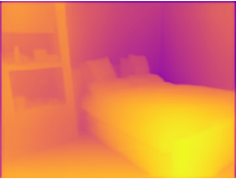} &
\includegraphics[width=0.23\linewidth]{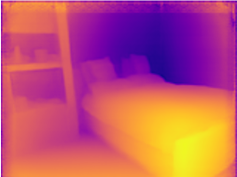} &
\includegraphics[width=0.23\linewidth]{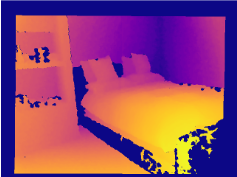} 
\\
Input & NeWCRFs \cite{newcrf} & Ours & GT
\end{tabular}
\caption{We observe that depth boundaries in state-of-the-art \cite{newcrf} align well with object boundaries, but the depth label is often incorrect. Note the confusion for the middle pillow in the first row and the bed in the second row. We propose skip attention module for fusing long range context from the decoder into the encoder features, which successfully mitigates the discrepancy.}
\vspace{-2.2em}
\label{fig:intro}
\end{center}
\end{figure}

%The encoder takes an RGB image as an input and produces feature maps representing the image at various resolutions. Due to the inherently local nature of a convolution kernel, early stage feature maps have higher resolution but a local receptive field. On the other hand, dense estimation tasks like MDE critically rely on a high receptive field as well as rich high-frequency resolution information for a more accurate estimation. 
Monocular Depth Estimation (MDE) is a well-studied topic in computer vision. State-of-the-art (\SOTA) techniques for MDE are based on encoder-decoder style Convolutional Neural Network (CNN) architectures~\cite{bts, eigen, dorn, yinetal, naderi, lee}. Due to the inherently local nature of a convolution kernel, early-stage feature maps have higher resolution but lack a global receptive field. The feature pyramidal-based decoder mitigates the issue by fusing low-resolution, semantically rich decoder features with the higher resolution but semantically weaker encoder features via a top-down path-way and lateral connections called skip connections \cite{fpn}.
Inline with the recent success of transformers, many latest works have used a self-attention based architectures for MDE~\cite{adabins, transdepth, newcrf}. Self-attention increases the receptive field and allows to capture long-range dependencies in feature maps. Practically, it is challenging to use self-attention for high-resolution feature maps due to memory and computational constraints. %Hence, Spatial Reduction Attention (SRA) \cite{pvt} and Window-based Attention (WBA) \cite{swin} are two popular approaches for efficiently computing self-attention at the initial stages of the encoder. SRA successfully increases the receptive field of the initial encoding stages but sacrifices information with high-frequency resolution. Window-based attention, on the other hand, limits its receptive field to a window without compromising the resolution information. 
Hence, the current \SOTA \cite{newcrf} uses window based attention using Swin transformer-based encoder backbone \cite{swin} to improve efficiency. 
%Other techniques such as Spatial Reduction Attention (SRA) \cite{pvt} have also been explored to improve tractability.

We observe that \SOTA \cite{adabins, newcrf} techniques are highly accurate in aligning depth edges with the object boundaries. However, there exists a confusion in giving the depth label to a pixel (c.f. \cref{fig:intro}). We posit this due to the inability of current techniques to effectively fuse high-resolution local features from the encoder and global contextual features from the decoder. Typically such a fusion is achieved through a skip connection module implementing feature concatenation followed by a convolution operation. Weights of the convolution kernels are highly localized, which restricts the flow of semantic information from long ranges affecting the ability of the model to predict the correct depth label for a pixel. 
% Yet, there are certain objects that are missing from depth map as shown in \ref{fig:intro}.
%
% Yet, the output depth map seems over-segmented as shown in \ref{fig:intro}. 
   %which restricts the flow of semantic information from long ranges,% 
To mitigate the constraint, we introduce a skip-attention module (SAM) that helps integrate information using window-based cross-attention. SAM calculates self-similarity between pixel queries based on decoder features and their corresponding neighbors from the encoder features in a predefined window to attend to and aggregate information at a longer range. We implement the overall architecture as a pixel query refinement problem. We use the coarsest feature map from the encoder with maximum global information to initialize pixel queries using a Pixel Query Initialiser Module. The pixel queries are then refined with the help of a SAM module to finer scales. 

Recent MDE techniques \cite{adabins} formulate the problem as a classification-regression one, in which the depth is predicted by a linear combination of bin centers discretized over the depth range. The bin centers are predicted adaptively per image, allowing the network to concentrate on the depth range regions that are more likely to occur in the scene of the input image. A vision transformer that aggregates global information from the output of another encoder-decoder-based transformer model is typically used to generate the bin centers. Since we pose MDE as a pixel query refinement problem starting from the coarsest resolution, we propose a lightweight Bin Center Module~(BCP) that predicts bin centers based on the initial pixel queries. This is more efficient than decoding features and then attending again in current \SOTA \cite{adabins}. The proposed design also helps embed the depth information into the initial pixel queries via direct ground truth supervision.
% in contrast to the loss propagated from the finest resolution feature query. 
%Hence, we propose a lightweight Bin Center Module~(BCP) that predicts bin centers based on the initial pixel queries. %Through experiments described later in the manuscript, we validate the significance of the proposed design.

% To regularize the learning problems, recent MDE techniques \cite{adabins}
% pose MDE as a classification-regression problem where the depth is predicted by a linear combination of bin centers discreticized over the depth range. The bin centers are predcted adaptive;ly per image that allows the netwrosk to adaptively focus on the regions of the depth range which are more probable to occur in the scene of the input image. 

% \textbf{Our second hypothesis for the over-segmentation of the depth map is due to the use of finer resolution for the depth bin prediction.} To regularize the learning problems, recent MDE techniques \cite{adabins} adaptively focus on the regions of the depth range which are more probable to occur in the scene of the input image. \textbf{However, they use a full transformer encoder-decoder backbone as a black box, and ends up using highest resolution feature maps for the bin prediction. This makes it harder to collate full-resolution information for globally optimal bin prediction. Instead, we introduce a Bin Center Predictor (BCP) Module that predicts image adaptive depth bin centers using coarsest level encoder features for the task.} We validate the significance of proposed change through our experiments described later in the manuscript.
 
\myfirstpara{Contributions}
The specific contributions of this work are as follows:
\begin{enumerate*}[label=\textbf{(\arabic*)}]
\itemsep0em 
\item We propose a novel strategy for predicting depth using a single image by viewing it as a pixel query refinement problem. 
\item We introduce a Skip Attention Module (SAM) that uses a window-based cross-attention module to refine pixel queries from the decoder feature maps for cross-attending to higher resolution encoder features. 
\item We present a Bin Center Predictor (BCP) Module that estimates bin centers adaptively per image using the global information from the coarsest-level feature maps. This helps to provide direct supervision to initial pixel queries from ground truth depth, leading to better query embedding.
\item We combine the novel design elements in an encoder-decoder framework comprised of a vision transformer backbone. The proposed architecture called PixelFormer achieves state-of-the-art (\SOTA) performance on indoor NYUV2 and outdoor KITTI datasets, improving the current \SOTA by 5.3\% and 3.9\%, in terms of absolute relative error and square relative error, respectively. Additionally, PixelFormer improves the generalization performance by 9.4\% over \SOTA on the SUNRGBD dataset in terms of absolute relative error.
\end{enumerate*}

\begin{figure*}[t]
	\begin{center}
		\includegraphics[width=0.99\linewidth]{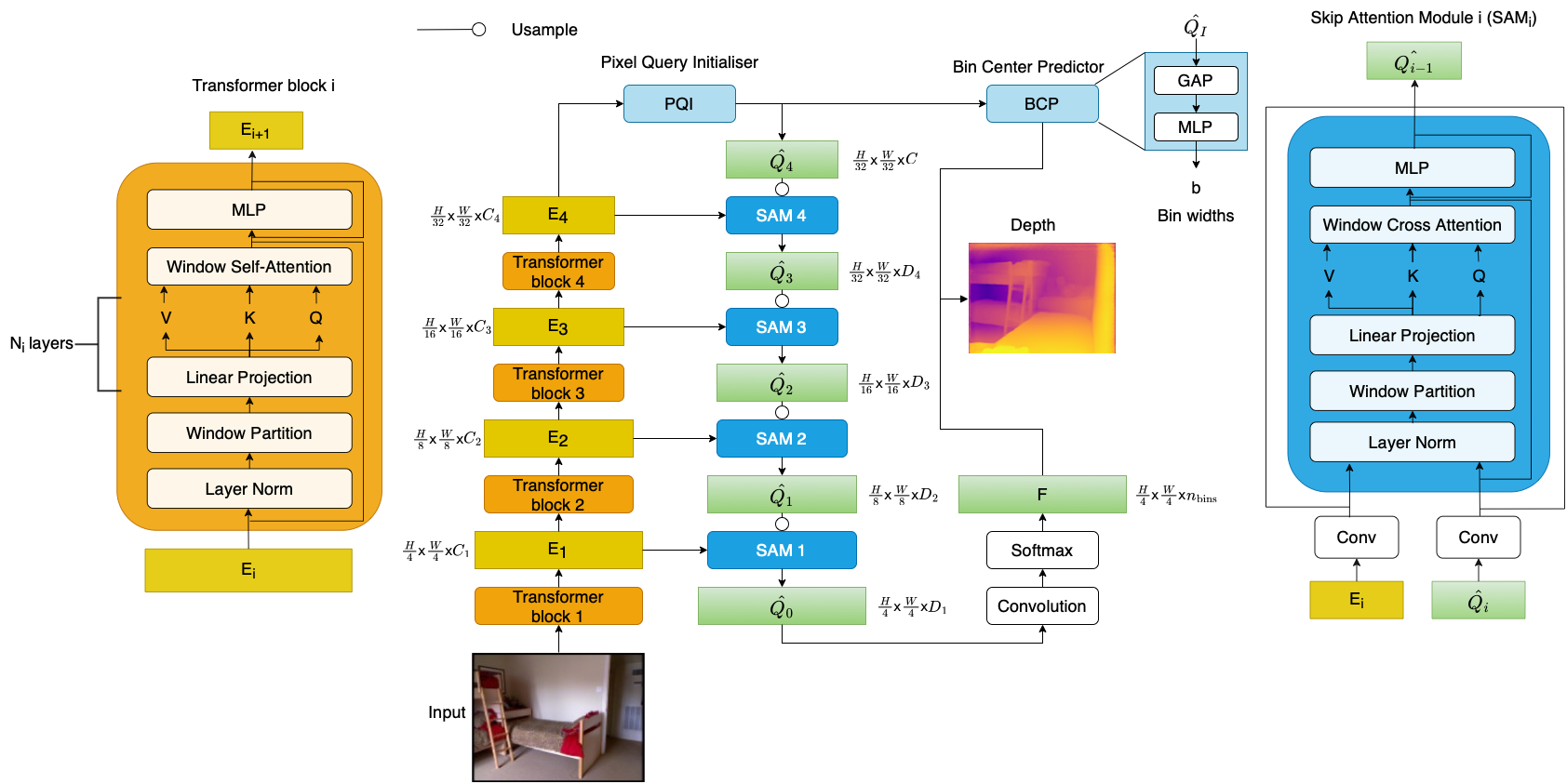}
	\end{center}
\vspace{-0.5em}
	\caption{\textbf{Detailed Architecture of our proposed approach PixelFormer :} Given an input image, a vision transformer-based encoder first extracts the multiscale feature maps. The feature map with the coarsest resolution ($E_4$) is given as input to the PQI module. The PQI module produces initial pixel queries that are given as input to the BCP module that produces the bin widths. The initial pixel queries are then refined to higher resolutions using the SAM modules. Finally, a convolution operation followed by a \emph{softmax} is applied to get probability distribution per pixel over the bin centers. }
	\label{fig:arch}
\end{figure*}

\section{Related Works}
\myfirstpara{CNN based MDE Techniques}
%
% Estimating depth using a single image is a long-standing problem, with the earlier works employing an encoder-decoder-based CNN. 
Eigen \etal \cite{eigen} first utilized CNN to predict depth from a single image by integrating global and local information. Song \etal \cite{song} have proposed a Laplacian pyramid-based model, and CLIFFNet \cite{cliffnet} a multi-scale convolutional fusion architecture to generate high-quality depth prediction. Yin \etal \cite{yinetal} introduced a geometric constraint named \emph{virtual normal}, and Naderi \etal \cite{naderi} proposed similarity between the RGB image and the corresponding depth map at the geometric edges to regularize predicted depth. Lee \etal \cite{lee} enforce a model to learn structural information about the scene by learning the relationship between image patches close to each other. Whereas, Patil \etal \cite{patil} exploit coplanar pixels to improve the predicted depth.

\mypara{Transformer based MDE Models}
Recent works have used Vision Transformer (ViT) architecture to improve the receptive field of a CNN in lower layers. Ranftl \etal \cite{dpt} uses a CNN to extract feature maps at ${\left(\frac{1}{16}\right)}^\text{th}$ resolution, which are passed to a vision transformer for global information aggregation. Bhat \etal \cite{adabins} uses a CNN-based encoder-decoder backbone and a ViT model to predict adaptive bins and pixel-level depths. NeWCRFs \cite{newcrf} uses Swin Transformer backbone \cite{swin} with CRFs at multiple scales. %In addition to using a swin transformer-based backbone inn our model, we introduce a window-based cross attention transformer to fuse the decoder feature queries and encoder features.

\mypara{MDE as Classification Vs Regression Task}
Modeling MDE as a regression problem leads to suboptimal solutions and faces convergence issues. Huan \etal \cite{dorn} first introduced the depth prediction task as a classification-regression problem solved by a CNN-based classification network in which the depth is predicted as a linear combination of bin centers discretized over the depth range. Recently, Bhat \etal \cite{adabins} proposed the prediction of bin centers adaptively per image using a ViT transformer on top of a transformer-based encoder-decoder backbone. In this work, we propose a single encoder-decoder backbone with a lightweight BCP module to predict bin centers using the coarsest resolution encoder feature maps.

\mypara{Skip Connections}
Skip connections were introduced by UNet \cite{unet} to forward high-resolution information from the encoder to the decoder via feature fusion. However, a naive fusion of early encoder and late decoder information is hindered by their semantic gap \cite{hybrid}. MultiResUnet \cite{MultiResUNet} replaces simplistic skip connection with a series of residual blocks to alleviate the semantic gap. Attention U-Net \cite{attentionunet} suppresses irrelevant regions in an input image while highlighting salient features useful for a specific task before the feature fusion. SANet \cite{Wen_2020_CVPR} uses attention to complete the point cloud at the decoder stage by injecting information using the encoder. In this work, we use skip attention to retrieve high-resolution details from encoder features using global contextual queries based on the decoder features. 
% fuse the local detail preserving encoder features with semantically strong decoder features using a vision transformer for MDE. 
%Note, we introduce SAM for a dense estimation task in contrast to a point cloud completion task \cite{Wen_2020_CVPR} where the input size is restricted to $\sim$2000 input points. Additionally, Point cloud completion has sparse neighborhood supervision for effective feature fusion, making it a simpler problem.

% SANet \cite{Wen_2020_CVPR} uses skip attention for point cloud completion to effectively exploit the local structure details of incomplete point clouds in the encoder stage during the inference of missing parts in the decoder stage. 

\section{Proposed Methodology}
%In this section, we provide an overview of the MDE problem and our proposed solution, including the architectural details and the training loss.

\myfirstpara{Problem Definition}
Following \cite{adabins, dorn}, we model MDE as a classification-regression task. Given an input image $I$, the network predicts bin widths, $b$, that discretize continuous depth range into an $n_\text{bins}$ number of intervals. The bins are predicted adaptively for each image. The final $n_\text{bins}$ dimensional probability vector is treated as the weight vector, and the depth, $d_i$, at a pixel $i$, is computed as a linear combination of the probability scores at the pixel with the predicted per-image bin-centers.

%\subsection{Proposed Architecture}

\mypara{Architecture Overview}
%
%In this work, we view MDE as a query refinement problem where the pixel-wise depth embeddings are refined in a multiscale fashion. To effectively gather resolution information from the encoder features, the decoder features, or the pixel queries in our formulation, query the encoder features. This strategy is applied hierarchically in a multiscale fashion to refine the pixel queries to incorporate the high-resolution pixel-level depth embeddings necessary for a depth estimation task. 
%
The input image $I$ is first fed to a Swin Transformer \cite{swin} that uses multiple layers of window-based self-attention to extract the feature maps representing the image at resolution scale $\{\frac{1}{4}, \frac{1}{8}, \frac{1}{16}, \frac{1}{32}\}$ \wrt $I$. The feature maps have a global receptive field due to the inherent nature of a ViT backbone. The feature map at ${\left(\frac{1}{32}\right)}^\text{th}$ scale is then fed to the proposed Pixel Query Initialiser (PQI) module. The PQI module aggregates the entire scene information using multi-scale global average pooling to initialize \emph{pixel queries}. The pixel queries are hierarchically refined with the encoder feature maps using the proposed \emph{Skip Attention Module} (SAM) deployed at various stages to predict per-pixel probability distribution over the bin centers. The initialized pixel queries are also sent to the \emph{Bin Center Predictor} (BCP) proposed in this work. BCP predicts bin centers adaptively per image using global average pooling followed by MLP layers. \cref{fig:arch} gives a pictorial description.
%The final depth for a pixel is estimated by a linear combination of the bin centers weighted by the probability distribution as in \cite{adabins}.

\mypara{Pixel Query Initialiser (PQI)}
The Pixel Query Initialiser (PQI) module aggregates the global information of the scene into each pixel-level embedding. The image feature map with the coarsest resolution, which contains the most essential details in the scene, is fed as an input to the PQI module. Given an input feature map of size $\frac{H}{32} \times \frac{W}{32} \times C_4$, the PQI module uses pyramid spatial pooling (PSP) \cite{psp} with an adaptive global pooling at scales 1, 2, 3, and 6. The feature maps are then upsampled to ${\frac{1}{32}}^{th}$ scale and concatenated. A convolution operation is then performed to integrate the global information effectively, as in \cite{newcrf}, to get initial pixel queries $Q_I$ of size $\frac{H}{32} \times \frac{W}{32} \times C$, where $C$ = 512.

\mypara{Bin Center Predictor (BCP)}
Previous works \cite{adabins} have used a vision transformer (ViT) to predict bin centers that discretize the image depth into a fixed number of intervals. ViT divides the image feature map into $16 \times 16$ patches and uses self-attention layers to exchange information among the patches. The first embedding is passed through an MLP head to predict the bin centers. Instead of decoding the feature map to high resolution and then using ViT, we propose to use the initial pixel queries to predict the bin centers. Apart from being more efficient, the proposed design helps in embedding the depth information into the pixel queries via direct ground truth supervision. 
Our BCP module consists of a simple Global Average Pooling followed by an MLP layer to predict the bin widths $b$ of dimension $n_\text{bins}$. Here, $n_\text{bins}$ denotes the number of adaptive bins per image. We use $n_\text{bins}=256$ for our model as suggested in \cite{adabins}. 
%The simple design and negligible computational overhead for our BCP module can be attributed to global information already aggregated into the pixel queries.
%
Given the pixel queries $Q_I$ of size $\frac{H}{32} \times \frac{W}{32} \times C$, we predict:
\begin{equation}
    b = \texttt{MLP}(\texttt{GAP}(Q))
\end{equation}
Finally, the bin centers for the input image are computed as:
\begin{align}
c \left( b_{i} \right) = d_\text{min} + \left( d_\text{max} - d_\text{min} \right) \bigg( \frac{b_i}{2} + \sum_{j=1}^{i-1} & b_j \bigg), \\ & i \in \{1,\ldots,n_\text{bins}\} \nonumber
\end{align}

\mypara{Skip Attention Module (SAM) Overview}
For a dense estimation task, coarse-level semantic features and fine-level details are both critical for an accurate estimation. Hence, similar to previous works \cite{fpn, newcrf}, we also use a bottom-down approach that starts with the lowest resolution feature map, upsample it, and injects the fine-level details from the encoder feature map at a particular scale using skip connections. 
%
%The popular encoder-decoder architectures merge the high-resolution information but semantically weak encoder features with semantically strong decoder features using skip connections. 
Typically skip connections use convolution operation after concatenating the encoder-decoder features.
%to introduce detailed information into the decoder features. %The problem with such an approach is the training bias that is introduced in the convolution kernel weights. It can be attributed to the fact that the fixed convolutional kernel weights are used to fuse the feature maps at all locations. To mitigate this bias, %
Unlike convolution operation to fuse the encoder-decoder features where the kernel weights are not adaptive as per the pixel location, we use the skip attention module (SAM), which uses self-similarity between the pixel queries and the corresponding encoder feature map, to effectively fuse the global-local features. 

\mypara{SAM Implementation}
Given a pixel query map $\hat{Q}_i$ and the corresponding encoder features $E_i$ for a particular scale $i$, we first perform a $3 \times 3$ convolution $E_i$ with $D_i$ channels on both $E_i$ and $\hat{Q}_i$ so that the number of channels of the pixel queries generated from the decoder features is the same as the number of channels in the encoder feature maps. Post the convolutional operation, a query matrix $Q$ is obtained from  $\hat{Q}_i$, and the key $K$ and value $V$ matrices are obtained from $E_i$ using the weights $W_q$, $W_k$ and $W_v$ implemented using MLP layers.
Since it's not computationally feasible for a query $q_i$ corresponding to the pixel query at location $i$ to attend to all the keys in the matrix $K$, we restrict the attention to a window as suggested for Swin Transformer \cite{swin}. $Q$, $K$, and $V$ matrices are first divided into windows of size $W \times W$. Similar to \cite{swin}, we use $W = 7$. Let $Q_w$, $K_w$, and $V_w$ be the query, key, and the value corresponding to the pixels in window $w$. We compute the output as follows:
\begin{equation*}
\text{Attention}(Q, K, V) = \text{Rearrange}(\text{Softmax}(Q_wK_w^T+B)V_w).
\end{equation*}
Here, $B$ denotes relative position bias. $B$ is a learnable matrix of size $w^2 \times w^2$, representing the relative position embedding corresponding to each query and key pair. The attention is computed for each window $w$ after which the rearrange operation places the windows as per their respective spatial location in Q.

To embed the information corresponding to the various depth ranges, each pixel query is divided into $H_i$ heads, and the attention operation is applied for each head. Post attention, per pixel depth embeddings are aggregated using MLP layers. The residual connections post attention and MLP layers are added for smooth gradient flows. To summarise, given $\hat{Q}_i$ and $E_i$ for pixel query and encoder at level $i$:
\begin{align*}
	\bar{Q}_i &= \text{LayerNorm}(\hat{Q}_i) \\
	\bar{E}_i &= \text{LayerNorm}(E_i) \\
	Q &= W_Q\bar{Q}_i, K = W_K\bar{E}_i, V = W_V\bar{E}_i \\
	\hat{Q}_{i-1} &= \text{MultiheadAttention}(Q, K, V)+\hat{Q}_{i} \\
	\hat{Q}_{i-1} &= \text{MLP}(\hat{Q}_{i-1}) +\hat{Q}_{i-1} \\
    \hat{Q}_{i-1} &= \text{MLP}(\hat{Q}_{i-1}) +\hat{Q}_{i} + E_i
\end{align*}
We have used $D_1, D_2, D_3, D_4$ = $\{$128, 256, 512, 1024$\}$ where $D_i$ corresponds to the number of channels in the convolutional kernel that is applied before the attention-based fusion of encoder and decoder features at stage $i$. The number of heads $H_1, H_2, H_3, H_4$ = $\{$4, 8, 16, 32$\}$ where $H_i$ represents the number of attention heads used in the SAM module at level $i$. More details can be seen in \cref{fig:arch}.% where we give a detailed architecture of our SAM module.

\mypara{Decoder Architecture}
As shown in \cref{fig:arch}, we start with an initial pixel query $\hat{Q}_I$ outputted from the PQI module. $\hat{Q}_I$ is upsampled to twice the resolution size using Pixel Shuffle \cite{pixelshuffle} and sent as input to the SAM module along with the corresponding encoder feature $E_4$. The initial pixel queries are refined to finer resolutions by attending to multiscale encoder feature maps at various resolutions through our proposed SAM module. For the given pixel query at level $\hat{Q}_i$ and the corresponding encoder feature $E_i$,
\begin{equation*}
    \hat{Q}_i = \text{SAM}(\text{Upsample}(\hat{Q}_{i+1},E_{i+1}) \quad
    i \in \{0, 1, 2, 3\}. 
\end{equation*}
Here, $\hat{Q}_4$ is same as $\hat{Q}_I$. A convolution operation is performed on $\hat{Q}_0$ to produce the final depth embedding $F$ of size $\frac{H}{32} \times \frac{W}{32} \times n_\text{bins}$. Finally, a pixel-wise \emph{softmax} operation is applied to obtain the per bin probability distribution $p_\text{bins}$:
\begin{equation}
    p_\text{bins} = \text{Softmax}(\text{Conv}(\hat{Q}_0))
\end{equation}
The final depth is predicted by the linear combination of the bin centers weighted by the probability values:
\begin{equation}
d_i = \sum_{k=1}^{n_\text{bins}} c \left( b_{k} \right) p_{ik},
\end{equation}
where $d_i$ is the predicted depth at pixel $i$, $c\left(b_{k}\right)$ is the $k^\text{th}$ bin center, $n_\text{bins}$ are the number of bins, and $p_{ik}$ is the probability for bin center $k$ for a pixel $i$.

\begin{figure*}
	\begin{center}
		\setlength\tabcolsep{2pt}
		\resizebox{0.9\textwidth}{!}{
			\renewcommand{\arraystretch}{1.1}
			\begin{tabular}{ccccc}
				\includegraphics[width=0.19\linewidth]{figures/rgb1.jpg} &
				\includegraphics[width=0.19\linewidth]{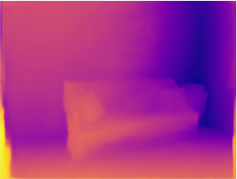} &
				\includegraphics[width=0.19\linewidth]{figures/crf1.png} &
				\includegraphics[width=0.19\linewidth]{figures/ours1.png} &
				\includegraphics[width=0.19\linewidth]{figures/gt1.png} 
				\\
				\includegraphics[width=0.19\linewidth]{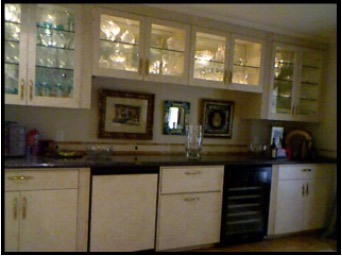} &
				\includegraphics[width=0.19\linewidth]{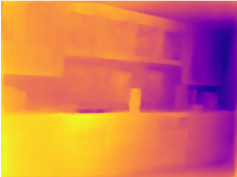} &
				\includegraphics[width=0.19\linewidth]{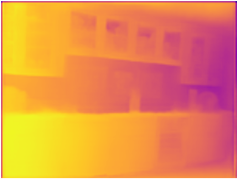} &
				\includegraphics[width=0.19\linewidth]{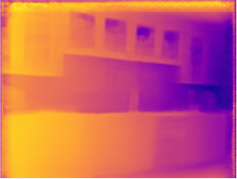} &
				\includegraphics[width=0.19\linewidth]{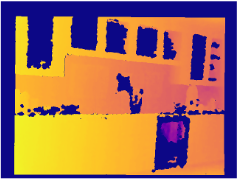} 
				\\
				\includegraphics[width=0.19\linewidth]{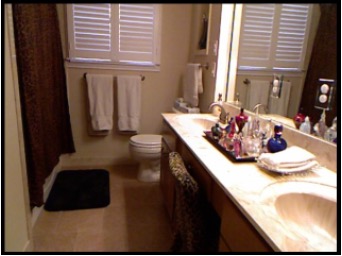} &
				\includegraphics[width=0.19\linewidth]{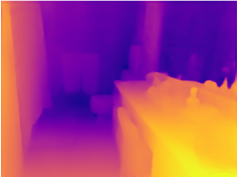} &
				\includegraphics[width=0.19\linewidth]{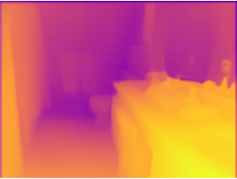} &
				\includegraphics[width=0.19\linewidth]{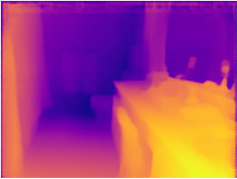} &
				\includegraphics[width=0.19\linewidth]{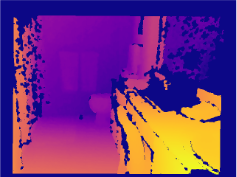} 
				\\
				\includegraphics[width=0.19\linewidth]{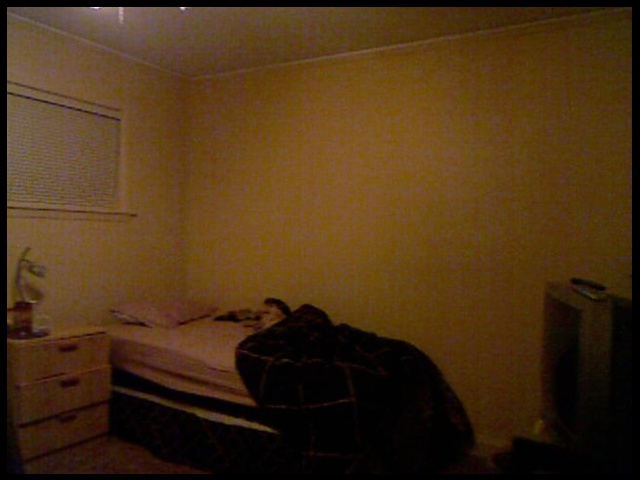} &
				\includegraphics[width=0.19\linewidth]{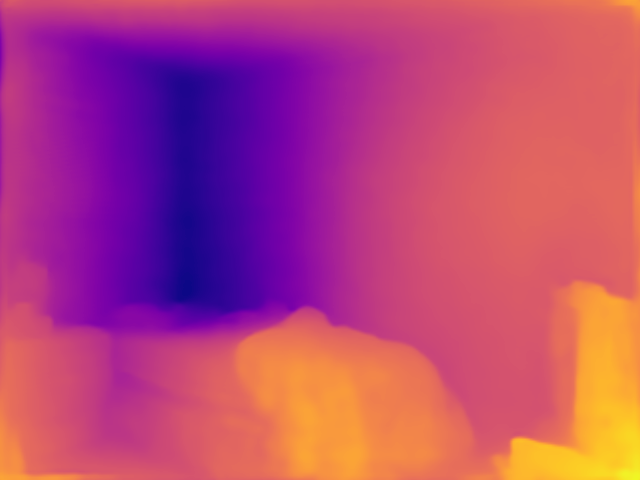} &
				\includegraphics[width=0.19\linewidth]{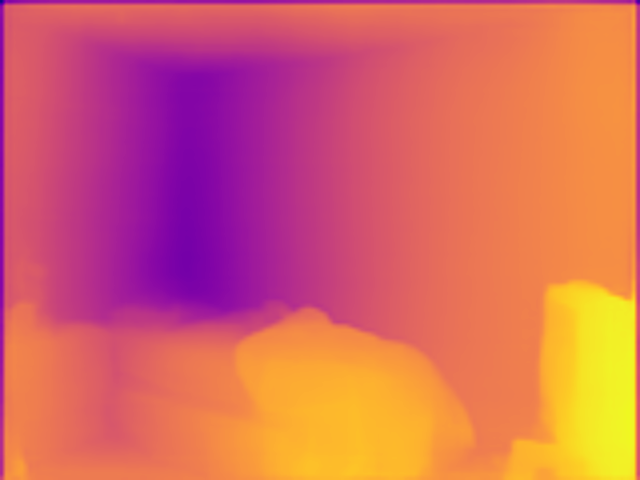} &
				\includegraphics[width=0.19\linewidth]{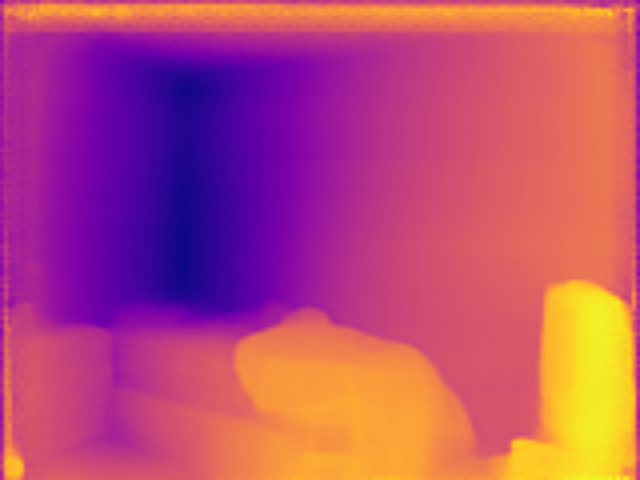} &
				\includegraphics[width=0.19\linewidth]{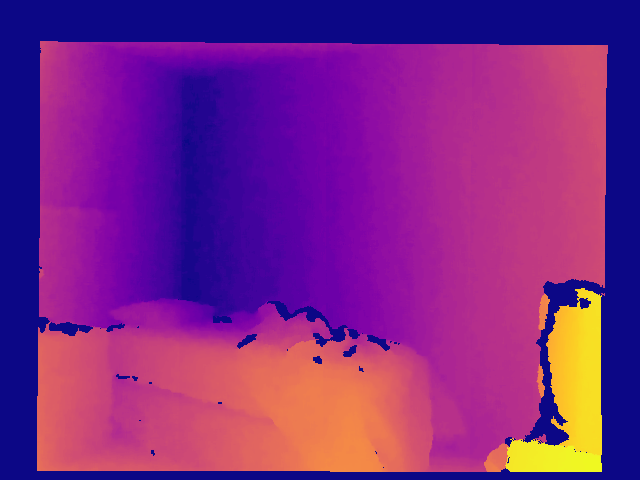} 
				\\
				Input & Adabins\cite{adabins} & NeWCRFs \cite{newcrf} & Ours & GT
		\end{tabular}}
		\caption{Qualitative comparison of our proposed method PixelFormer on the indoor dataset NYUV2 against Adabins and NeWCRFs.}
		\vspace{-1.5em}
		\label{fig:nyu}
	\end{center}
\end{figure*}

\mypara{Training loss}
Following previous works \cite{newcrf, adabins}, we use a scaled version of scaled version of the Scale-Invariant loss (SILog) \cite{eigen} to supervise our network. Given, the ground truth depth ($d_i^*$) and the predicted depth ($d_i$) at a pixel location $i$, first the logarithmic distance between $d_i$ and $d_i^*$ is calculated as: $g_{i}=\log(\hat{d}_{i})-\log(d_{i}^{*})$. The SIlog loss is then calculated as follows:
\begin{equation}
\mathcal{L}_\text{SILog} = \alpha \sqrt{\frac{1}{n} \sum_{i} g_{i}^{2}-\frac{\lambda}{n^{2}}\left(\sum_{i} g_{i}\right)^{2}}.
\end{equation}
Here, $n$ denotes the number of pixels in an image that have the ground truth values available. Following \cite{adabins}, we use $\lambda$ = 0.85 and $\alpha$ = 10 for all our experiments. % as the scaling parameters.

\section{Datasets and Evaluation}
\myfirstpara{NYU Depth V2}
NYUV2 \cite{nyu} is an indoor dataset containing 120K RGB and depth pairs with a size of $480 \times 640$ acquired as video sequences from 464 indoor scenes using a Microsoft Kinect. We follow the official training/testing split to evaluate our method, where 249 scenes comprising 50K images are used for training, and 654 images from 215 scenes are used for testing. We use the center cropping proposed by Eigen \etal \cite{eigen}, with the depth maps having an upper bound of 10 meters. Our network outputs depth prediction with a resolution of $120 \times 160$, which we upsample by $4\times$ to match the ground truth resolution during training and testing.

\mypara{KITTI Dataset} 
KITTI \cite{kitti} is an outdoor dataset that consists of stereo images and 3D scans from 61 scenes captured by multiple sensors mounted on top of a moving vehicle. The dataset contains an input RGB image with a $1241 \times 375$ pixels resolution, and the LIDAR scans correspond to it. We use the training/testing split defined by \cite{eigen} that consists of a subset of 26K left view images from the official Kitti dataset for the training and 697 test set images. To evaluate the test set, we use the crop defined by Garg \etal \cite{garg} with the depth maps having an upper bound of 80 meters. We use bilinear interpolation to upsample the prediction to match the ground truth image resolution.

\mypara{SUNRGB-D}
SUNRGB-D \cite{Song_2015_CVPR} is an indoor dataset collected using various sensors. It comprises 10335 real RGB-D images of room scenes. The training and testing sets contain 5285 and 5050 images, respectively. We use the official test set for evaluation purposes with an upper bound on the depth of 8 meters.

\mypara{Evaluation Metrics}
We use the standard metrics Average relative error (Abs Rel), Root mean squared error (RMSE), Average Log error ($\log_{10}$), Threshold Accuracy ($\delta_i$) at thresholds $\tau_i's = 1.25, 1.25^2, 1.25^3$ used in earlier works \cite{eigen, adabins, newcrf} to compare our method against state-of-the-art. For KITTI evaluation, we additionally use Square relative error (Sq Rel).
% %
% Given the predicted depth $d_p$, and the ground truth depth $d^*_p$ at a pixel $p$, and $n$ denoting the total number of pixels in an image, the error metrics are defined as: 
% \begin{itemize}[leftmargin=*]
% %
% \item Average relative error (Abs Rel):
% \[\frac{1}{n} \sum_{p = 1}^{n} \frac{\left|d_{p}-d^{*}_{p}\right|}{d^{*}_{p}} \]
% %
% \item Root mean squared error (RMSE):
% \[ \sqrt{\left.\frac{1}{n} \sum_{p = 1}^{n}\left(d_{p}-d^{*}_{p}\right)^{2}\right)} \]
% %
% \item Average Log error ($\log_{10}$): 
% \[ \frac{1}{n} \sum_{p = 1}^{n} \big| \log_{10}(d_{p})-\log_{10}(d^{*}_{p}) \big| \]
% %
% \item Threshold Accuracy (at threshold $\tau$): Percentage of $d_{p}$ such that $\max \left(\frac{d_{p}}{d^{*}_{p}}, \frac{d^{*}_{p}}{d_{p}}\right) < \tau$. We report at $\tau$ values $1.25$, $1.25^2$, and $1.25^3$.  
% %
% \item For KITTI evaluation, we additionally use,
% Square relative error (Sq Rel) 
% \[\frac{1}{n} \sum_{p = 1}^{n} \frac{||d_{p}-d^{*}_{p}||^2}{d^{*}_{p}}.\]
% %
% \end{itemize}

\section{Experiments}
\begin{table*}[t]
\centering
\setlength{\tabcolsep}{12pt}
\begin{tabular}{lccccccc}
\toprule
Method                      & Venue  &  Abs Rel{$\downarrow$} & RMSE{$\downarrow$} & {{$\log_{10}$}{$\downarrow$}} & {{$\delta_1$}{$\uparrow$}} & {{$\delta_2$}{$\uparrow$}} & {{$\delta_3$}{$\uparrow$}}  \\   
\midrule
Eigen \etal \cite{eigen}  & NIPS'14    & 0.158                   & 0.641                   & \multicolumn{1}{c}{-}     & 0.769                      & 0.950                      & 0.988                      \\
DORN \cite{dorn}                        & CVPR'18    & 0.115                   & 0.509                   & 0.051                     & 0.828                      & 0.965                      & 0.992                      \\
Yin \etal \cite{yinetal}    & ICCV'19    & 0.108                   & 0.416                   & 0.048                     & 0.872                      & 0.976                      & 0.994                      \\
BTS \cite{bts}                         & Arxiv'19 & 0.110                   & 0.392                   & 0.047                     & 0.885                      & 0.978                      & 0.994                      \\
DAV \cite{dav}                         & ECCV'20    & 0.108                   & 0.412                   & \multicolumn{1}{c}{--}    & 0.882                      & 0.980                      & 0.996                      \\     
TransDepth  \cite{transdepth}                & ICCV'21    & 0.106                   & 0.365                   & 0.045                     & 0.900                      & 0.983                      & 0.996                      \\
DPT* \cite{dpt}                        & ICCV'21    & 0.110                   & 0.367                   & 0.045                     & 0.904                      & 0.988                      & \textbf{0.998}                     \\
PackNet-SAN* \cite{packnetsan}               & CVPR'21    & 0.106                   & 0.393                   & \multicolumn{1}{c}{--}    & 0.892                      & 0.979                      & 0.995     \\           
Adabins \cite{adabins}                     & CVPR'21    & 0.103                   & 0.364                   & 0.044                     & 0.903                      & 0.984                      & \underline{0.997}                     \\
Naderi \etal \cite{naderi} & WACV'22    & 0.097                   & 0.444                   & 0.042                     & 0.897                      & 0.982                      & 0.996                      \\
Lee \etal \cite{lee}    & WACV'22    & 0.107                   & 0.373                   & 0.046                     & 0.893                      & 0.985                      & \underline{0.997}                      \\
P3Depth \cite{patil} & CVPR'22 & 0.104 & 0.356 &  0.043  & 0.898 & 0.981 & 0.996 \\
NeWCRFs \cite{newcrf}                      & CVPR'22    & \underline{0.095}                   & \underline{0.334}                   & \underline{0.041}                     & \underline{0.922}                      & \textbf{0.992}                      & \textbf{0.998}                      \\
\midrule
\textbf{PixelFormer (ours) }         &            & \textbf{0.090}                   & \textbf{0.322}                   & \textbf{0.039}                     & \textbf{0.929}                      & \underline{0.991}                      & \textbf{0.998}  \\
\bottomrule
\end{tabular}
\caption{Results on NYUV2 \cite{nyu} Dataset. The best results are in \textbf{bold} and second best are \underline{underlined}. “*” means using additional data for training.  $\uparrow$ means higher the better and $\downarrow$ means lower the better. An upper bound of 10 meters on the ground truth depth map is used for evaluation. All the numbers have been taken from the corresponding papers. We see an overall improvement against the \SOTA in terms of almost all the metrics used for evaluation.}
\label{tab:nyu}
\end{table*}

\begin{table*}[t]
	\centering
	\setlength{\tabcolsep}{8pt}
\begin{tabular}{lcccccccc}
\toprule
Method                      & Venue  &  Sq Rel{$\downarrow$} & Abs Rel{$\downarrow$} & RMSE{$\downarrow$} & {{$\log_{10}$}{$\downarrow$}} & {{$\delta_1$}{$\uparrow$}} & {{$\delta_2$}{$\uparrow$}} & {{$\delta_3$}{$\uparrow$}} \\                      
\midrule
Eigen \etal  \cite{eigen}           & NIPS'14              & 1.548                     & 0.203                         & 6.307                         & 0.282                                                & 0.702                        & 0.898                        & 0.967                        \\
Godard \etal \cite{Godard_2017_CVPR}            & CVPR'17              & 0.898                     & 0.114                         & 4.935                         & 0.206                                                & 0.861                        & 0.960                        & 0.976                        \\
Kuznietsov \etal \cite{Kuznietsov_2017_CVPR}        & CVPR'17              & 0.741                     & 0.113                         & 4.621                         & 0.189                                                & 0.862                        & 0.964                        & 0.986                        \\
Gan \etal \cite{gan}              & ECCV'18              & 0.666                     & 0.098                         & 3.933                         & 0.173                                                & 0.890                        & 0.984                        & 0.985                        \\
DORN \cite{dorn}                                   & CVPR'18              & 0.307                     & 0.072                         & 2.727                         & 0.120                                                & 0.932                        & 0.984                        & 0.994                        \\
Yin \etal   \cite{yinetal}           & ICCV'19              & -                         & 0.072                         & 3.258                         & 0.117                                                & 0.938                        & 0.990                        & \underline{0.998}                        \\
BTS \cite{bts}                                & Arxiv 19             & 0.245                     & 0.059                         & 2.756                         & 0.096                                                & 0.956                        & 0.993                        & \underline{0.998}                       \\
PackNet-SAN* \cite{packnetsan}                            & ICCV'21              & -                         & 0.062                         & 2.888                         & -                                                    & 0.955                        & -                            & -                            \\
TransDepth \cite{transdepth}                            & ICCV'21              & 0.252                     & 0.064                         & 2.755                         & 0.098                                                & 0.956                        & 0.994                        & 0.994                        \\
Adabins \cite{adabins}                                & CVPR'21              & 0.190                     & 0.058                         & 2.360                         & 0.088                                                & 0.964                        & \underline{0.995}                        & \textbf{0.999}                        \\
DPT* \cite{dpt}                                   & ICCV'21              & -                         & 0.060                         & 2.573                         & 0.092                                                & 0.959                        & \underline{0.995}                        & 0.996                        \\
Naderi \etal \cite{naderi}            & WACV'22              &      & 0.070 & 3.223 & 0.113                        & 0.944                        & 0.991                        & \underline{0.998}                        \\
NeWCRFs \cite{newcrf}                                 & CVPR'22              & \underline{0.155} & \underline{0.052}     & \underline{2.129}     & \underline{0.079}                           & \underline{0.974}   & \textbf{0.997}    & \textbf{0.999}    \\
\midrule
\textbf{PixelFormer (ours)} &  & \textbf{0.149}                     & \textbf{0.051}                          & \textbf{2.081}                          &  \textbf{0.077} & \textbf{0.976} & \textbf{0.997} & \textbf{0.999} \\
\bottomrule
\end{tabular}
\caption{Results on KITTI Eigen Split test set \cite{eigen}. The best results are in \textbf{bold} and second best are \underline{underlined}. ``*'' means using additional data for training. $\uparrow$ means higher the better and $\downarrow$ means lower the better. An upper bound of 80 meters on the ground truth depth map is used for evaluation.  All the numbers have been taken from the corresponding papers.  }
\label{tab:kitti}
\end{table*}
\begin{table*}[t]
	\centering
	\setlength{\tabcolsep}{8pt}
\begin{tabular}{lcccccccc}
\toprule
Method                      & Venue  &  Sq Rel{$\downarrow$} & Abs Rel{$\downarrow$} & RMSE{$\downarrow$} & {{$\log_{10}$}{$\downarrow$}} & {{$\delta_1$}{$\uparrow$}} & {{$\delta_2$}{$\uparrow$}} & {{$\delta_3$}{$\uparrow$}} \\                        
\midrule
Chen \etal \cite{chen}  & IJCAI'19 & -  & 0.166                  & 0.494                   & 0.071     & 0.757                      & 0.943                & \underline{0.984}                      \\
Yin \etal \cite{yinetal}  & ICCV'19 & -      & 0.183                   & 0.541                   & 0.082                     & 0.696                      & 0.912                      & 0.973                      \\
BTS \cite{bts}        & Arxiv'19 &    -              & 0.172                   & 0.515                   & 0.075                     & 0.740                      & 0.933                      & 0.980                    \\
Adabins \cite{adabins}   & CVPR'21 &     -                     & \underline{0.159}                   & \underline{0.476}                   & \underline{0.068}    & \underline{0.771}                      & \underline{0.944}                      & 0.983                      \\   
\midrule
\textbf{PixelFormer (ours) }   &      & \textbf{0.0915}  &   \textbf{0.144}       & \textbf{0.441}                   & \textbf{0.062}                      & \textbf{0.802}                      & \textbf{0.962}                      & \textbf{0.990}  \\
\bottomrule
\end{tabular}
\caption{Results on SUNRGB-D test set  without fine-tuning the models trained on NYUV2. The best results are in \textbf{bold} and second best are \underline{underlined}.  $\uparrow$ means higher the better and $\downarrow$ means lower the better. An upper bound of 8 meters on the ground truth depth map is used for evaluation. The numbers have been taken from the \cite{adabins}.}
\label{tab:sunrgbd}
\end{table*}

%To validate our proposed method, Pixelformer's effectiveness in predicting the depth of a single image, we conduct extensive experiments on both outdoor and indoor scenes. In this section, we describe the dataset used for evaluation, implementation details, evaluation metrics, and the results of our proposed approach that achieves state-of-the-art performance on MDE task.

\myfirstpara{Implementation Details}
The proposed method is implemented in Pytorch \cite{pytorch}. We use Adam optimizer \cite{adam} ($\beta's$= 0.9, 0.999), with a batch size of 8 and a weight decay of $10^{-2}$. We use 20 epochs for both KITTI and NYUV2 datasets, with an initial learning rate of $4 \times 10^{-5}$, which is decreased linearly to $4 \times 10^{-6}$ across the training iterations. Our model takes 30 minutes per epoch using 4 NVIDIA A100 GPUs. We use various data augmentation techniques like random rotation, horizontal flipping, changing the image brightness, and Cut Depth \cite{cutdepth}. We use the pre-trained weights of Swin-L \cite{swin} to initialize our encoder backbone. We follow a similar test protocol as in \cite{adabins, newcrf}, and output final depth values by averaging the predicted depth for the original image and its mirror image. 

%\subsection{Comparison with the State-of-the-art}

\mypara{Results on NYUV2}
\cref{tab:nyu} and \cref{fig:nyu} show the quantitative and qualitative results, respectively, on the indoor dataset NYUV2 using our approach, named PixelFormer. Following the test protocol of \cite{dpt, packnetsan}, and without additional training data, our method improves the Absolute relative error by 5.3\% over \SOTA. The performance gain is significant considering the saturated performance of the dataset in recent years. We see an improvement of 9.6\% and 3.5\% over the recently proposed methods \cite{patil} and NeWCRFs, respectively, in terms of the RMSE error. We see in \cref{fig:nyu} that PixelFormer produces more accurate depth maps than Adabins and NeWCRFs, which can be attributed to the proposed SAM module, which allows capturing long-range dependencies. In contrast to other approaches, PixelFormer can estimate depth maps corresponding to missing objects, as shown in the third row of \cref{fig:nyu}.

% predict better depth boundaries. It can be attributed to the fact that the fine level details are injected into the decoder features via cross attention which is pixel adaptive.

\begin{figure*}[t]
\includegraphics[width=\linewidth]{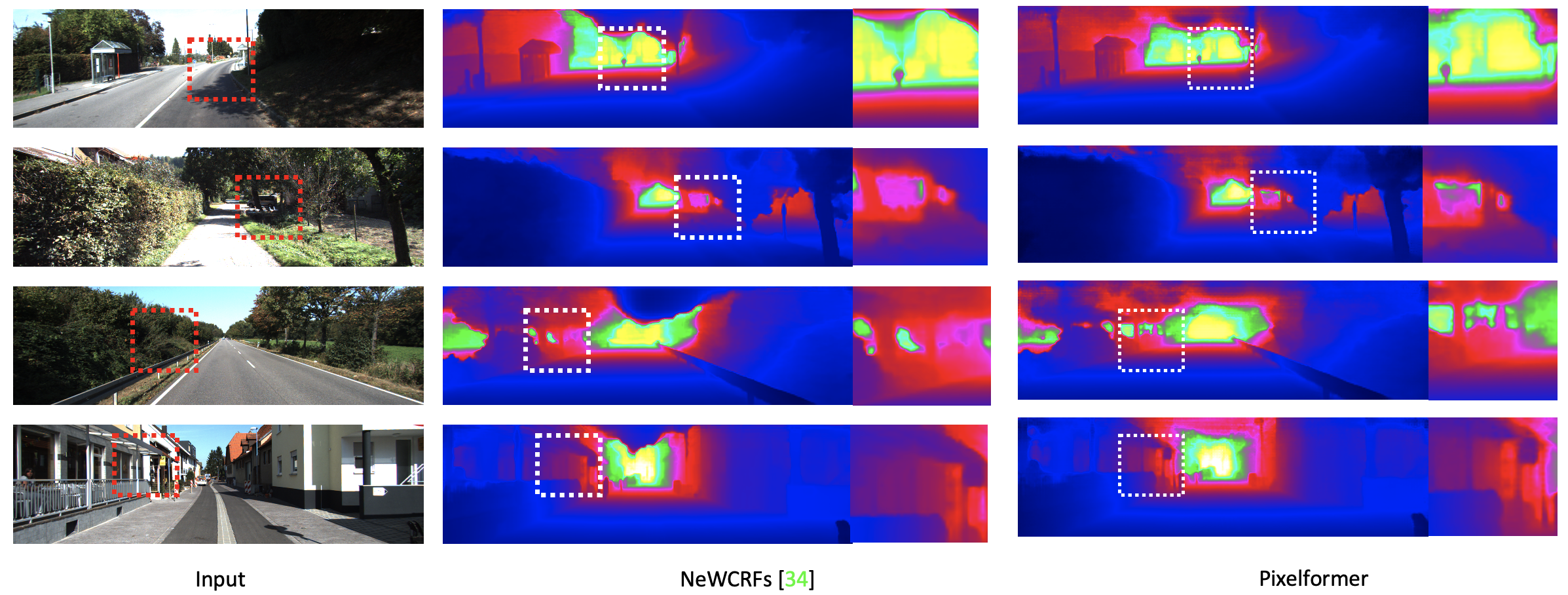}
\caption{Qualitative comparison of previous state-of-the-art method NeWCRFs on the outdoor dataset KITTI.}
\label{fig:kitti}
\end{figure*}

\mypara{Results on KITTI}
\cref{tab:kitti} and \cref{fig:kitti} show the quantitative and qualitative results, respectively, on the outdoor dataset KITTI. We see an overall improvement of 3.9\% and 2.3\% in terms of Sq. Rel and RMSE respectively against the \SOTA NeWCRFs \cite{newcrf} on the KITTI Eigen Split. We also compare our method against the previous \SOTA approaches on the official KITTI test set. Currently, we rank $1^{st}$ on the official benchmark\footnote{\url{http://www.cvlibs.net/datasets/kitti/eval_depth.php?benchmark=depth_prediction}} against the previous peer-reviewed approaches with an improvement of 2.5\% in terms of Abs Rel and 1.1\% in terms SILog against NeWCRFs. 
As shown in the top row of \cref{fig:kitti}, PixelFormer can precisely estimate depth maps for far-sighted road objects.

% From Figure \ref{fig:kitti}, it is quite evident that Pixelformer can predict better depth boundaries against the previous works for outdoor scenes as well, proving the efficacy of content adaptive fusion.

\mypara{Results on SUNRGB-D}
Following \cite{adabins}, we analyze our network's generalization performance by evaluating the model performance on test SUNRGB-D without finetuning the model on the NYUV2 dataset. As shown in \cref{tab:sunrgbd}, PixelFormer outperforms Adabins by 9.4\% and 7.4\% in terms of Abs Rel and RMSE, respectively. Thus demonstrating the effectiveness of pixel-adaptive global local fusion for out-of-distribution input images.

\section{Ablation Study}
%In this section, we conduct the ablation experiments to demonstrate the efficacy our proposed approach Pixelformer.

\myfirstpara{Efficacy of Skip Attention Module}
\begin{table}[t]
\centering
\setlength{\tabcolsep}{8pt}
\begin{tabular}{lccc}
\toprule
Method &  Abs Rel{$\downarrow$} & Sq Rel{$\downarrow$} & {{$\delta_1$}{$\uparrow$}} \\
\midrule
Add-Conv  &  0.0602 & 0.190 & 0.964  \\
Cat-Conv & 0.0613 & 0.192 &  0.964 \\
\midrule
Decoder-Ours (SAM) & \textbf{0.0578} & \textbf{0.182}  & \textbf{0.967}  \\
\bottomrule
\end{tabular}
\caption{Ablation experiment to demonstrate the efficacy of SAM module on KITTI Eigen Split using Swin-T as the encoder. $\uparrow$ means higher the better and $\downarrow$ means lower the better. The best results are in \textbf{bold} and second best are \underline{underlined}. }
\label{tab:abl_sam}
\end{table}
\cref{tab:abl_sam} demonstrates the effectiveness of our proposed SAM module against other baseline convolution-based alternatives \emph{Add-Conv}  and \emph{Cat-Conv} to combine the encoder and decoder features at a particular scale. Add-Conv fuses the encoder-decoder features by pixel-wise addition followed by a convolution operation. Cat-Conv concatenates the encoder and decoder features \wrt the channel dimension, followed by a convolution operation. We see that the addition-based approach outperforms the concatenation-based approach by a small margin. However, using our SAM module outperforms  Add-Conv by 4.0\% in terms of Abs Rel and 4.2\% in terms of Sq. Rel. This validates the contribution of the proposed SAM module. 

\begin{table}[t]
\centering
\setlength{\tabcolsep}{12pt}
\begin{tabular}{lccc}
\toprule
Method &  Abs Rel{$\downarrow$} & Sq Rel{$\downarrow$} & {{$\delta_1$}{$\uparrow$}} \\
\midrule
mViT-Last  & 0.0596 & 0.190  & 0.964\\
mViT-First & 0.0584 & 0.185 &  0.966 \\
\midrule
Ours (BCP) & \textbf{0.0578} & \textbf{0.183}  & \textbf{0.967} \\
\bottomrule
\end{tabular}
\caption{Ablation experiment to demonstrate the usefulness of embedding the depth information into the initial pixel queries on KITTI Eigen Split using Swin-T as the encoder. $\uparrow$ means higher the better and $\downarrow$ means lower the better. The best results are in \textbf{bold} and second best are \underline{underlined}.}
\label{tab:abl_bcp}
\end{table}

\mypara{Effectiveness of embedding the depth information into the pixel queries}
We experiment to showcase the usefulness of using the initial pixel queries to predict the bin centers. We compare our proposed design that predicts the bin centers using the initial pixel queries against using a vision transformer to predict the bin centers as in \cite{adabins}. mViT-Last converts the feature map of the highest resolution ($F$ shown in \cref{fig:arch}) into 16$\times$16 patches and integrates the information in the first patch using multiple self-attention layers ($L = 4$). The first patch embedding is passed through MLP layers to predict the bin centers. Similarly, mViT-First predicts bin centers by passing the initial pixel queries to a ViT. We use patch size = 1 for mViT-First for a fair comparison. \cref{tab:abl_bcp} shows that both mVit-First and our approach outperform mVit-Last by 2.0\% and 3.0\%, respectively, in terms of Abs Rel, indicating that embedding depth information into the initial pixel queries via direct loss supervision helps predict better depth estimates. mVit-First does not give any further benefit to predicting bin centers since global information is already aggregated into the initial pixel queries via the PQI module.

% \subsection{HyperParameters}
% \subsubsection{Number of Bins}
% \subsubsection{Number of Heads}

\section{Conclusion}
This work presents PixelFormer, a novel encoder-decoder strategy for Monocular Depth Estimation that poses the problem as a pixel query refinement problem. The global initial pixel queries predicted by the Pixel Query Initialiser module are refined to a higher resolution by querying the multiscale encoder features at various resolutions through the proposed Skip Attention Module. Unlike convolution-based skip connections, the module can fuse decoder features with long-range dependency, leading to more accurate depth labels. Our proposed Bin Center Prediction module helps constrain the network with depth information embedded into the initial pixel queries through direct loss supervision. Through extensive experiments, we showcase that PixelFormer improves state-of-the-art performance on the indoor dataset NYUV2 and outdoor dataset KITTI by 5.3\% and 3.9\%, respectively, along with an improved generalization performance by 9.4\% on the indoor SUNRGBD dataset. In the future, we will try to apply our content adaptive fusion using SAM to other dense estimation tasks like semantic segmentation.

\mypara{Acknowledgement} This work has been partly supported by the funding received from DST through the IMPRINT program (IMP/2019/000250). We acknowledge National Supercomputing Mission (NSM) for providing computing resources of `PARAM Siddhi-AI', under National PARAM Supercomputing Facility, CDAC Pune, and supported by the Ministry of Electronics and Information Technology and Department of Science and Technology, Government of India.

{\small
\bibliographystyle{ieee_fullname}
\bibliography{egbib}
}

\end{document}